\documentclass[lettersize,journal]{IEEEtran}
\usepackage{amsmath,amsfonts}
\usepackage{algorithmic}
\usepackage{algorithm}
\usepackage{array}
\usepackage[caption=false,font=normalsize,labelfont=sf,textfont=sf]{subfig}
\usepackage{textcomp}
\usepackage{stfloats}
\usepackage{url}
\usepackage{verbatim}
\usepackage{graphicx}
\usepackage{cite}

\usepackage{booktabs}
\usepackage{multirow}
\usepackage{caption} 
\usepackage{subcaption} 
\usepackage{multirow, multicol}
\usepackage{color}
 \usepackage[colorlinks]{hyperref}

\usepackage{hyperref}

\begin{document}

\title{ExFusion: Efficient Transformer Training via Multi-Experts Fusion}
\author{Jiacheng Ruan, Daize Dong, Xiaoye Qu, Tong Zhu, Ting Liu, Yuzhuo Fu, Yu Cheng and Suncheng Xiang,~\IEEEmembership{Member,~IEEE}
\thanks{Corresponding authors: Yuzhuo Fu, Yu Cheng and Suncheng Xiang.}
\thanks{Jiacheng Ruan, Ting Liu, Yuzhuo Fu are with the School of Electronic Information and Electrical Engineering, Shanghai Jiao Tong University, No 800, Dongchuan Road, Minhang District, Shanghai, China, 200240  (email: jackchenruan@sjtu.edu.cn, louisa\_liu@sjtu.edu.cn, yzfu@sjtu.edu.cn).}
\thanks{Daize Dong is with the Shanghai Artificial Intelligence Laboratory, Shanghai, China, 200240  (email: dongdaize.d@pjlab.org.cn).}
\thanks{Xiaoye Qu is with the Huazhong University of Science and Technology, Hubei, China, 430074  (email: xiaoye@hust.edu.cn).}
\thanks{Tong Zhu is with the Soochow University, Jiangsu, China, 215031 (email: tzhu1997@outlook.com).}
\thanks{Yu Cheng is with the Chinese University of Hong Kong, Hongkong, China, 999077 (email: chengyu05@gmail.com).}
\thanks{Suncheng Xiang is with the School of Biomedical Engineering, Shanghai Jiao Tong University, No 800, Dongchuan Road, Minhang District, Shanghai, China, 200240 (email:xiangsuncheng17@sjtu.edu.cn).}
}


\markboth{Journal of \LaTeX\ Class Files,~Vol.~14, No.~8, August~2021}%
{Shell \MakeLowercase{\textit{et al.}}: A Sample Article Using IEEEtran.cls for IEEE Journals}


\maketitle

\begin{abstract}
Mixture-of-Experts (MoE) models substantially improve performance by increasing the capacity of dense architectures. However, directly training MoE models requires considerable computational resources and introduces extra overhead in parameter storage and deployment. Therefore, it is critical to develop an approach that leverages the multi-expert capability of MoE to enhance performance while incurring minimal additional cost. To this end, we propose a novel pre-training approach, termed ExFusion, which improves the efficiency of Transformer training through multi-expert fusion. Specifically, during the initialization phase, ExFusion upcycles the feed-forward network (FFN) of the Transformer into a multi-expert configuration, where each expert is assigned a weight for later parameter fusion. During training, these weights allow multiple experts to be fused into a single unified expert equivalent to the original FFN, which is subsequently used for forward computation. As a result, ExFusion introduces multi-expert characteristics into the training process while incurring only marginal computational cost compared to standard dense training. After training, the learned weights are used to integrate multi-experts into a single unified expert, thereby eliminating additional overhead in storage and deployment. Extensive experiments on a variety of computer vision and natural language processing tasks demonstrate the effectiveness of the proposed method. The code will be released publicly upon acceptance.
\end{abstract}


\section{Introduction}

\begin{figure}[!t]
  \centering
  \begin{minipage}[t]{0.42\textwidth}
    \centering
    \includegraphics[width=\textwidth]{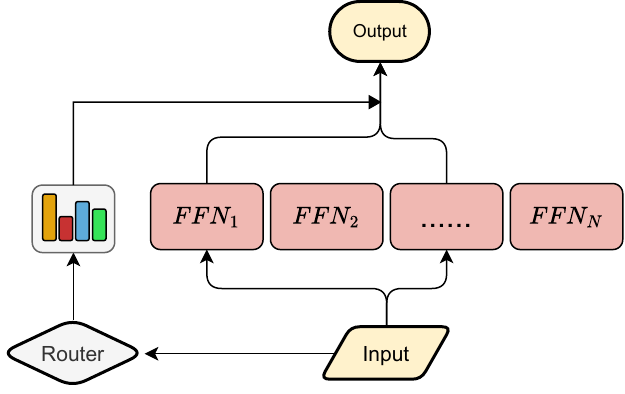} 
    \caption*{\textbf{(a)} Vanilla Top-$k$ MoE.}
  \end{minipage}
  \hfill
  \begin{minipage}[t]{0.42\textwidth}
    \centering
    \includegraphics[width=\textwidth]{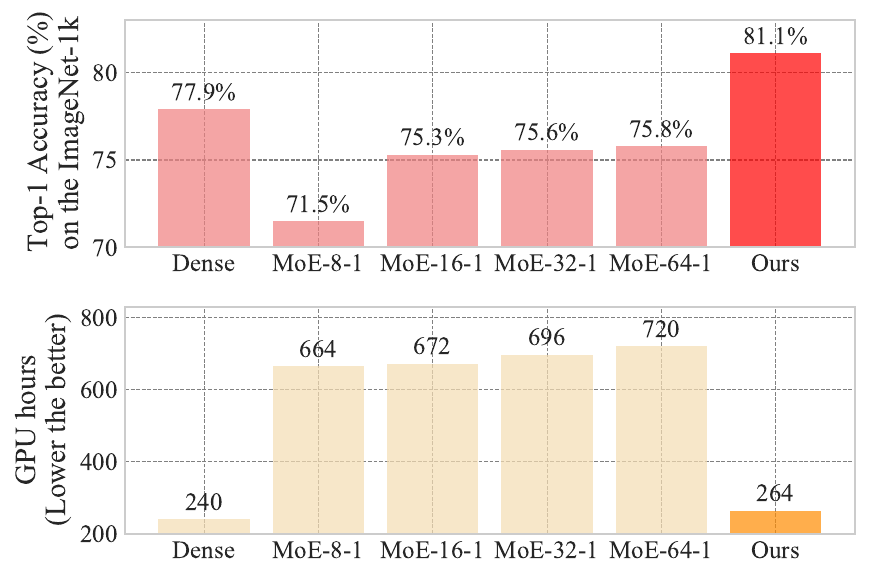} 
    \caption*{\textbf{(b)} Performance \& A100 GPU hours.}
  \end{minipage}
  \caption{
      \textbf{(a)} Illustration of the vanilla Top-$k$ MoE. Given an input, the MoE layer employs a router to identify and select the top $k$ most pertinent experts from $N$. In this paper, we refer to this configuration of MoE as MoE-$N$-$k$.
      \textbf{(b)} Comparison of the performance and training GPU hours among the baseline dense model (ViT-B/16), vanilla MoE models, and our ExFusion. All models are pre-trained on the ImageNet-1k dataset and share the same number of activated parameters.
  }
  \label{fig:head}
\end{figure}

\IEEEPARstart{T}{ransformer}-based models have rapidly progressed in computer vision (CV) \cite{vit,deit}, demonstrating strong performance across tasks including image classification \cite{pvt,cvt,Dynamicvit,swin}, object detection \cite{DETR,DeformableDETR,dynamicDETR,vitdet}, and semantic segmentation \cite{maskformer,segformer,SETR}. Recently, large-scale models have catalyzed increasing interest in scaling the parameterization of Transformer-based architectures. In this context, the Mixture-of-Experts (MoE) architecture has attracted substantial attention \cite{lstmmoe,GLaMmoe,switchTRM,softmoe,llama-moe-2023}. As illustrated in Figure \ref{fig:head}(a), MoE substitutes the feed-forward network (FFN) layer in a standard Transformer with an MoE layer, which comprises multiple FFN modules (each serving as an expert) and a trainable router. The router selectively activates a small subset of experts to process each input. This multi-expert design enhances representational capacity and promotes learning through expert specialization and collaboration. Consequently, MoE models offer notable advantages for large-scale datasets, particularly in settings characterized by complex and diverse tasks.


However, despite their advantages, MoE models have several inherent limitations. Firstly, although MoE models often outperform their dense counterparts, these gains are offset by substantially increased parameter counts and higher training costs.
For example, SwinV2-MoE-S (128 experts) \cite{tutel}, trained on a large-scale dataset (e.g., ImageNet-22k \cite{imagenet22k}), surpasses SwinV2-S by roughly 1\% on the ImageNet-1k benchmark \cite{imagenet1k}. However, this improvement requires nearly a 30× increase in parameters and additional training data.
Secondly, deploying MoE models for inference also demands the accommodation of a significant number of parameters within computation devices, posing serious challenges for numerous researchers \cite{moedeploy}. 
Lastly, as illustrated in Figure \ref{fig:head}(b), we conduct a series of experiments with the ViT-B/16 and its corresponding MoE models\footnote{Tutel package \cite{tutel} is employed to implement MoE models, which is an advanced package specifically designed for constructing high-performance MoE models.} on the ImageNet-1k dataset. Under standard training recipes\footnote{Training Transformer-based models on the ImageNet-1k dataset typically involves setting a batch size of 1,024 and training for 300 epochs. For more detailed training strategies, please refer to Section \ref{sec:imnet-1k}.}, MoE models are markedly less efficient. For instance, on ImageNet-1k, the MoE-64-1 ViT-B/16, trained with 3x GPU hours compared to ViT-B/16, still exhibits a 2.1\% performance degradation. Consistent with prior findings, MoE models can underperform under standard training settings and constrained resources \cite{scalinglaw,v-moe}.


In this paper, we pose a challenging question: \textit{Is it feasible to fully leverage the advantages of MoE while avoiding a substantial increase in overhead (e.g., parameter size, computational cost, and training time)?} Inspired by the averaging method in ensemble learning \cite{ensemblelearningsurvey}, we propose a simple yet effective approach that fuses the parameters of multiple experts. This strategy retains the representational benefits of the MoE architecture while reducing overhead. To this end, we present Multi-Experts Fusion (ExFusion), a training method for Transformer-based models in CV and natural language processing (NLP). Specifically, ExFusion assigns a fusion weight to each expert and uses these weights to combine expert parameters into a single set of parameters, yielding a unified expert that processes the input. Depending on the weight-update mechanism, ExFusion has three variants: Static-Weight, Dynamic-Weight, and Memory-Bank ExFusion. The first two variants remove the router and instead use fixed or learnable weights to control the fusion process. In contrast, the third variant retains the router and introduces an additional memory bank. The memory bank stores the router-derived weights at each iteration and updates them using momentum. After training, we use the final fusion weights to merge multiple experts into a single expert and retain only the parameters of the fused expert. This design ensures that the parameter count and computational complexity of the resulting model match those of the corresponding dense model.

To comprehensively demonstrate the efficacy of ExFusion, we conduct extensive experiments across CV and NLP tasks. For the CV tasks, we evaluate four Transformer-based models (ViT \cite{vit}, DeiT \cite{deit}, PVT \cite{pvt}, Swin \cite{swin}) and perform pre-training on the ImageNet-1k dataset. After pre-training, we fine-tune the resulting models for semantic segmentation and object detection. In the realm of NLP, we opt for the T5 model for various experiments. Consistent with the CV protocol, we first pre-train T5 on the large-scale C4 dataset \cite{t5-c4} and then fine-tune it on the GLUE benchmark to evaluate its generalization across diverse NLP tasks. These experiments not only confirm the effectiveness of our approach but also showcase its broad applicability and flexibility across different types of tasks.

The main contributions are summarized as follows:
 \begin{itemize}
\item For the first time, we present a novel parameter-fusion method for efficient training of Transformer models, called ExFusion. Our method integrates multiple experts in MoE models during both training and post-training via associated fusion weights.
\item We systematically investigate alternative weighting schemes and propose three variants of ExFusion: static weighting, dynamic weighting, and a memory-bank mechanism. These variants provide flexible and efficient control over parameter fusion.
\item We conduct extensive experiments on both CV and NLP tasks. The results demonstrate that ExFusion achieves superior performance without increasing parameter count or computational cost. ExFusion provides a foundation for future research on efficient training frameworks across diverse tasks and domains.
\end{itemize}

In the remainder of this paper, we first review related work on Transformer-based models, mixture-of-experts (MoE), ensemble learning and other efficient training and merging
methods in Section \ref{sec2}. In Section \ref{sec3}, we present the architecture and learning procedure of the proposed pre-training approach, referred to as ExFusion. Extensive evaluations against state-of-the-art methods and comprehensive analyses of the proposed approach are provided in Section \ref{sec4}. We conclude the paper and discuss future work in Section \ref{sec5}.

\section{Related Work}
\label{sec2}

\subsection{Transformer-Based Models}
The Transformer is an architecture composed of stacked self-attention modules and FFN layers, originally introduced for machine translation \cite{nlptrm}. Building on the Transformer architecture, numerous language models have been proposed \cite{BERT, t5-c4, gpt1}, demonstrating strong performance in NLP. The success of the Transformer has subsequently extended to CV. Vision Transformer (ViT) \cite{vit} applies the Transformer to image classification by tokenizing images into patches, achieving competitive results. DeiT \cite{deit} introduces token-based distillation to transfer knowledge from a teacher model in a data-efficient, data-driven manner, leading to improved performance on visual tasks. PVT \cite{pvt} further proposes a pyramidal hierarchy that progressively reduces spatial resolution, and introduces spatially sparse attention patterns to better support dense prediction tasks. To improve computational efficiency, Swin Transformer \cite{swin} employs shifted local-window attention, enabling cross-window information exchange while reducing computational cost. CSWin \cite{Cswin} introduces cross-shaped window self-attention that computes horizontal and vertical stripe-wise attention in parallel, improving efficiency. DAT \cite{DAT} proposes a dual aggregation strategy that alternates spatial window self-attention and channel-wise self-attention across successive blocks and incorporates an adaptive interaction module and a spatial-gated feed-forward network, thereby facilitating the fusion of spatial and channel features. NAT \cite{NAT} emulates convolutional inductive bias by restricting attention to a local neighborhood around each feature; specifically, it adopts a convolution-inspired, query-centric formulation that selects key–value pairs within each query’s local neighborhood to compute attention. GroupMixFormer \cite{GroupMixFormer} introduces Group-Mix Attention, which jointly models correlations across multiple granularities (token-to-token, token-to-group, and group-to-group) and feature groups; it serves as an efficient alternative to standard self-attention and can achieve strong visual performance with fewer parameters. In addition, other studies \cite{liu2023efficientvit,yun2024shvit,vasu2023fastvit,nottebaum2025lowformer} have further advanced Transformer-based architectures in the field of CV. 

In this paper, we propose an approach that improves existing Transformer-based models via a training-centric strategy rather than designing new self-attention mechanisms. By leveraging a MoE architecture and a weighted fusion scheme, our ExFusion improves performance with slightly overhead.

\subsection{Mixture-of-Experts}


The Mixture-of-Experts (MoE) layer introduces structured sparsity and is often used to replace the feed-forward network (FFN) in Transformer architectures to expand model capacity. In machine translation, Gshard \cite{GShardmoe} replaces FFN layers with MoE and achieves substantial gains in translation quality. Switch Transformer \cite{switchTRM} scales MoE to the trillion-parameter regime for language model pretraining, enabling unprecedented sparsity while improving performance. V-MoE \cite{v-moe} incorporates MoE layers into ViT and is pretrained on large-scale datasets (e.g., JFT-300M \cite{JFT-300M} and JFT-3B \cite{JFT-3B}) for approximately 20K TPUv3-core-days, achieving strong image classification performance. Built on Swin Transformer V2, SwinV2-MoE \cite{tutel} is pretrained on ImageNet-22K \cite{imagenet22k} using 128 A100 GPUs, yielding improvements on downstream tasks. Moreover, subsequent studies have advanced MoE across natural language processing \cite{expertchoicerouter-moe,moebert,meomoe,flanmoe,mixtralmoe,deepseekmoe}, visual recognition \cite{moedg,softmoe,ewamoe}, multimodal learning \cite{Vlmo-moe,scalingVLMOE,chen2023eve,Moe-llava}, and other domains \cite{tmm_moe1,tmm_moe2,tmm_moe3,tmm_moe4}.

However, training and inference for MoE models often require substantial computational resources (Table \ref{fig:head}), which can be prohibitive for the broader community. In this paper, we propose ExFusion, which significantly reduces the training and inference overhead of MoE while preserving its capacity-scaling benefits.

\subsection{Ensemble Learning}

Ensemble learning \cite{zhou2012ensemble,ensemblelearningsurvey} is a fundamental component of machine learning, widely applied across various domains, utilizing methods such as bagging \cite{bagging,breiman2001random}, boosting \cite{xgboost,augboost}, and stacking \cite{stacking}. Among these strategies, the averaging method is particularly prevalent in deep learning \cite{fisher_avg, stochastic_wa, fuse_pretrain, xiang2023learning}, combining multiple weak models to produce a more robust model through weighted averaging. For instance, Singh et al. \cite{optimal_transport} employ optimal transport theory to average the weights of trained models, achieving minimal accuracy loss. Ainsworth et al. \cite{git_basin} align model weights through permutation symmetries, thereby improving the effectiveness of the averaging approach. In downstream tasks, Model Soups \cite{model_soup} demonstrates that averaging the parameters of multiple fine-tuned models enhances prediction accuracy without incurring additional inference time costs. Meanwhile, RegMean \cite{regmean} explores the averaging of models fine-tuned on diverse tasks and introduces an innovative and effective method. Furthermore, methods such as Expert Weights Averaging (EWA) and Weight-Ensembling MoE (WEMoE)~\cite{tang2024merging} improve performance without adding extra inference overhead by utilizing weight averaging or expert fusion techniques.

In contrast to the aforementioned methods, we investigate the training-time weight averaging strategy. Our ExFusion method employs weighted fusion of parameters from multiple experts during training, followed by post-training weight averaging to compress the model and enhance efficiency in both training and inference.

\begin{figure*}[!t]
    \centering
    \includegraphics[width=0.95\textwidth]{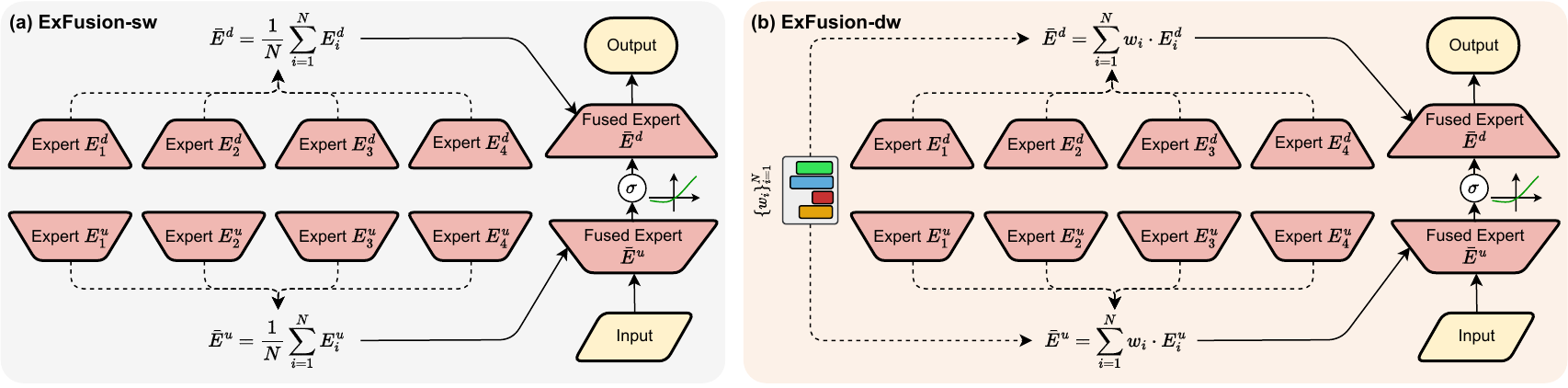}
    \caption{The overall architecture of ExFusion-sw and ExFusion-dw. The number of experts is set to four for demonstration. The dashed lines represent the multi-experts fusion process.}
    \label{fig:swdw}
\end{figure*}

\section{Preliminaries}
\label{sec3}

\subsection{Transformer}

In this section, we first review the general Transformer model. Specifically, after the input data undergoes embedding processing, it can be represented as $x \in \mathbb{R}^{L \times D}$, where $L$ denotes the sequence length and $D$ denotes the dimension of the embedding. The sequence is then fed into multiple Transformer blocks for feature extraction, each block consisting of a multi-head self-attention (MHSA) module followed by a feed-forward network (FFN) layer. Within each head, the self-attention is represented as:
\begin{gather}
    q = W^q(x), \quad k = W^k(x), \quad v = W^v(x), \\
    h_i = \sum_{j=1}^N \frac{\textup{Sim}(q_i,k_j)}{\sum_{j=1}^N \textup{Sim}(q_i,k_j)}v_j,
\end{gather}
where $W^q, W^k, W^v \in \mathbb{R}^{D \times D}$ denote linear layers, and $\textup{Sim}(\cdot,\cdot)$ represents the similarity function. Subsequently, the intermediate features $h \in \mathbb{R}^{L\times D}$ obtained from the MHSA module are fed into an FFN layer to produce the output $y \in \mathbb{R}^{L\times D}$, which can be expressed as follows:
\begin{equation}
y = W^{d}\left(\sigma\left(W^{u}(h)\right)\right),
\end{equation}
where $W^{u} \in \mathbb{R}^{D \times eD}$ and $W^{d} \in \mathbb{R}^{eD \times D}$ represent linear layers, where $e$ is the expansion factor, typically set to 4. $\sigma$ represents the activation function, commonly chosen as GELU \cite{gelu-act}.

\subsection{MoE Layer}
\label{sec:pre-moe-layer}

The MoE layer typically consists of $N$ expert networks $\{E_1,E_2,\cdots,E_n\}$ and a router $G$. Each expert is an independent FFN layer, while the router contains a trainable linear layer designed to determine the importance of each expert. Given an input $x\in\mathbb{R}^{L\times D}$, the importance is represented as follows:

\begin{equation}
G(x) = \textup{\textup{Softmax}}(W^g(x)),
\end{equation}
where $W_g \in \mathbb{R}^{D \times N}$ denotes the linear layer in the router. $G(x) \in \mathbb{R}^{L\times N}$ represents the importance distribution of each expert being selected for a given input. Typically, the output of the MoE layer is the sum of the outputs from $k$ selected experts ($k \le N$), which can be represented as follows:

\begin{equation}
y = \sum_{i \in \mathcal{K}} G_i(x) \cdot E_i(x),
\end{equation}
where $G_i(x)$ is the weight allocated to the $i$-th expert, $E_i(x)$ is the output of the $i$-th expert, and $\mathcal{K}$ is the set of selected top-$k$ indices determined by $G(x)$.

\section{Multi-Experts Fusion (ExFusion)}


While Mixture-of-Experts (MoE) models have the potential to deliver outstanding performance by expanding model capacity, they also introduce considerable challenges, particularly in terms of escalating the requirements for training, storage, and deployment resources.
This raises a critical question: \textit{Could we harness the benefits of MoE during training yet retain a parameter count and computational complexity comparable to dense models, all without a significant rise in training expenses?}
Our investigation approaches this query from an ensemble learning standpoint, applying an averaging technique that dynamically combines the experts, which is called Multi-Experts Fusion (ExFusion).
In the following sections, we first elaborate on the motivation behind ExFusion in Sec. \ref{Motivation}.
We then respectively introduce three variants of ExFusion in Sec. \ref{sec:ExFusionsa}, \ref{sec:ExFusiondw} and \ref{ExFusionmb}, which are distinguished by the type of fusion weights.


\subsection{Motivation}
\label{Motivation}

We begin by introducing the averaging method derived from ensemble learning. Consider a scenario where there are $k$ independent weak models, each with prediction errors that are independent and identically distributed (i.i.d.), and can be modeled as random variables $Y_i$ with variance $\text{Var}(Y_i) = \sigma^2$.
The ensemble of model predictions $Y$ can then be calculated as the average of predictions from all weak models, represented by $Y = \frac{1}{k} \sum_{i=1}^{k} Y_i$. The variance of prediction $Y$ is:

\begin{equation}
\text{Var}(Y) = \left(\frac{1}{k^2}\right) \sum_{i=1}^{k} \text{Var}(Y_i) = \frac{\sigma^2}{k}.
\label{eq:average}
\end{equation}
Equation \ref{eq:average} demonstrates that the overall variance of the model significantly decreases after applying the averaging method. Under linear approximation, the prediction error of a single model can be regarded as $\epsilon_i = b + \xi_i$ comprising a bias term $b$ and zero-mean noise $\xi_i$. When averaging multiple experts uniformly, the bias term remains unchanged, while the noise variance is reduced to a fraction of its original value $\sigma^2/k$, thereby lowering the overall generalization error without significantly increasing the bias.

This highlights the effectiveness of the averaging method in leveraging the diversity among different models to reduce variance and enhance the overall stability of the model, thereby improving the model's performance.
Drawing inspiration from this observation, we shift our focus to a new perspective on the MoE, treating each expert as a weak model that plays a role in a weighted averaging process.
Specifically, consider an MoE layer containing $N$ experts, represented as $\{E_1,E_2,\cdots,E_N\}$ with a corresponding set of weights $\{w_1,w_2,\cdots,w_N\}$.
We posit that each expert processes the input $x$ independently, with their outputs subsequently merged through a weighted fusion mechanism.
This process can be viewed as analogous to initially blending the experts according to their respective weights, thereby creating a singular fused expert $\bar{E}$, which then processes the input.
Consequently, this suggests the following implication:

\begin{equation}
    Y = \sum_{i=1}^N w_i \cdot E_i(x) = \left[\sum_{i=1}^N w_i \cdot E_i\right](x) = \bar{E}(x).
\end{equation}

Different from Sec. \ref{sec:pre-moe-layer}, \(E_i\) represents an individual linear layer instead of an entire FFN layer.
This design is inspired by Equation \ref{eq:average} and is targeted at boosting model efficacy by diminishing the cumulative variance.
Building on this concept, we introduce ExFusion, a method crafted to optimize the training of Transformer models through the weighted fusion of multiple experts.
More specifically, we develop three variants step by step: the first is ExFusion-sw, which utilizes static weights; the second is ExFusion-dw, employing learnable weights; and the third is ExFusion-mb, which generates weights based on data-dependent information.

To sum up, among the methods for improving generalization through weight averaging, SWA~\cite{stochastic_wa} enhances generalization by averaging weights of multiple models sampled from the later stages of a single SGD training trajectory, Model Soups~\cite{model_soup} integrates the advantages of multiple models fine-tuned from the same pre-trained initialization with different hyperparameters via weight averaging, and EWA~\cite{ewamoe} improves generalization by replacing some FFNs of Vision Transformers with MoEs (Mixture-of-Experts) using random uniform token partition during training and averaging expert weights to convert MoEs back to FFNs for inference, without additional inference cost. Different from the above methods, our ExFusion not only fuses the multiple experts of MoE via averaging after training, but also performs the fusion operation during the training phase, reducing the overhead in both training and inference stages.

\begin{figure*}[!t]
    \centering
    \includegraphics[width=0.95\textwidth]{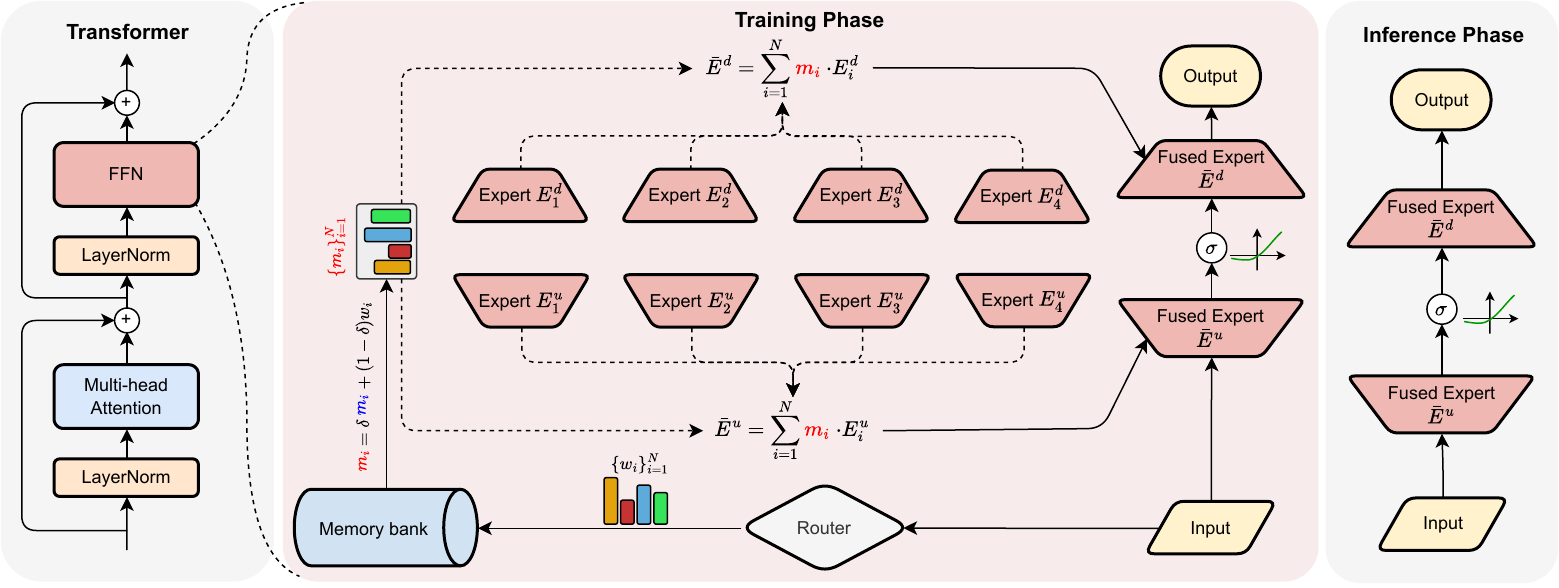}
    \caption{The overall architecture of ExFusion-mb. During training, the router and memory bank control the importance of experts depending on the input data. After training, the final weights in the memory bank is employed to fuse experts. The number of experts is set to four for demonstration. The dashed lines represent the multi-experts fusion process.}
    \label{fig:main}
\end{figure*}

\subsection{Static-Weights ExFusion (ExFusion-sw)}
\label{sec:ExFusionsa}
ExFusion-sw combines the parameters of multiple experts via a simple uniform averaging, as illustrated in Figure \ref{fig:swdw} (a).
To be specific, before training, two sets of experts $\{E_1^u, E_2^u, \cdots, E_N^u\}$ and $\{E_1^d, E_2^d, \cdots, E_N^d\}$ are initialized to replace two linear layers in the FFN layer, $W^u$ and $W^d$.
Each expert is then assigned a static weight of $\frac{1}{N}$ for subsequent fusion.
During the training process, we fuse each expert set into a single expert and use the fused expert to process the input $x$, resulting in output $y$.
This process can be represented as:
\begin{equation}
    \begin{aligned}
        \bar{E}^{u} &= \frac{1}{N} \sum_{i=1}^{N} E_{i}^{u}, \quad \bar{E}^{d} = \frac{1}{N} \sum_{i=1}^{N} E_{i}^{d}, \\
        y &= \bar{E}^{d} \left( \sigma\left(\bar{E}^{u}(x)\right) \right),
    \end{aligned}
\label{eq:ExFusionsw}
\end{equation}
where $\bar{E}^u$ and $\bar{E}^d$ are fused experts. Upon training completion, we continue to employ the fixed weight $\frac{1}{N}$ to fuse multiple experts, retaining only the parameters of $\bar{E}^u$ and $\bar{E}^d$ for storage. The advantage of this approach lies in its ability to leverage multiple experts to learn more nuanced representations, while also harnessing the concept of ensemble learning to amalgamate these experts. As a result, our approach maintains the parameter count and computational complexity as same as the dense model, while providing an extended model capacity as the MoE model without compromising the model's efficiency.



\subsection{Dynamic-Weights ExFusion (ExFusion-dw)}
\label{sec:ExFusiondw}
Based on the static fusion strategy, we further introduce a dynamic learning mechanism to adjust the weight corresponding to each expert, as shown in Figure \ref{fig:swdw} (b).
For two sets of experts, we introduce a shared set containing $N$ learnable weights, denoted as $\mathbf{W}=\{w_1, w_2, \cdots, w_N\}$.
During training, we utilize $\textbf{W}$ to represent the importance of each expert and perform a weighted fusion:

\begin{equation}
    \begin{aligned}
        \bar{E}^u = \sum_{i=1}^N w_i \cdot E_i^u, \quad \bar{E}^d = \sum_{i=1}^N w_i \cdot E_i^d.
    \end{aligned}
\label{eq:ExFusiondw}  
\end{equation}

After training, we utilize the optimized weights to fuse the multiple experts, preserving solely the parameters of the unified expert.
The merit of our dynamic weighting scheme lies in its capacity to autonomously re-calibrate the weights allocated to the experts, whose distribution signifies the contribution of each expert to the model, thereby enhancing the model's flexibility and adaptability under different conditions.


\subsection{Memory-Bank ExFusion (ExFusion-mb)}
\label{ExFusionmb}

In MoE models, the router stands out as a crucial part of identifying the importance of experts for each input.
Despite the capability of adaptively adjusting the importance of experts, ExFusion-dw neglects the differences of the input data.
To address this oversight, we introduce the router and memory bank into ExFusion-mb to account for the influence of input data on the importance of experts, as illustrated in Figure \ref{fig:main}.
Specifically, before training, a linear layer $G$ is introduced to serve as the router, along with a memory bank $\mathbf{M} = \{m_1, m_2, \cdots, m_N\}$ whose parameters are initialized as zeros to store the updates of weights.
Throughout the training process, the input $x$ is processed via the router to derive the weight set $\mathbf{W} = \{w_1, w_2, \cdots, w_N\}$ for the current iteration, and the memory bank $\mathbf{M}$ is updated in accordance with momentum $\delta$:
\begin{equation}
\begin{aligned}
    &\mathbf{W} = \{w_1, w_2, \cdots, w_N\} = \textup{Softmax}(G(x)), \\
    &\mathbf{M} = \{\delta m_i + (1-\delta) w_i\}_{i=1}^N.
\end{aligned}
\label{eq:ExFusionmb-1}  
\end{equation}

The weights in $\mathbf{M}$ is then employed to fuse multi-experts:
\begin{equation}
\begin{aligned}
    \bar{E}^u = \sum_{i=1}^N m_i \cdot E_i^u, \quad \bar{E}^d = \sum_{i=1}^N m_i \cdot E_i^d.
\end{aligned}
\label{eq:ExFusionmb-2}  
\end{equation}

Similarly, the final set of weights $\mathbf{M}$ is used for the weighted fusion of multi-experts after training.
Subsequently, both $G$ and $\mathbf{M}$ are discarded to reduce the parameter storage.
As a result, ExFusion-mb ensures steady updates of expert weights while leveraging data-driven insights, effectively combining efficiency with performance.

\section{Experiments}
\label{sec4}

To verify the effectiveness of ExFusion, we initially perform pre-training on the ImageNet-1k dataset \cite{imagenet1k}.
Subsequently, the fused models are fine-tuned for object detection using the COCO dataset~\cite{cocodataset} and semantic segmentation with the ADE20K dataset~\cite{ade20k}.
Furthermore, in illustration of the versatility of our method, we also conduct language modeling experiments on the C4 dataset \cite{t5-c4} for NLP tasks, followed by fine-tuning on the GLUE benchmark \cite{gluebenchmark}.
Models that are trained using ExFusion-mb are marked with the dagger symbol ($\dagger$). Additionally, we fix the number of experts to 4 and set the momentum of the memory bank to 0.95 for ExFusion-mb.
All experiments are conducted with 8 NVIDIA A100 GPUs. 

\begin{table}[!t]
\centering
\setlength{\tabcolsep}{1.25pt} 
\caption{Comparison results of Dense ViT-B/16, the corresponding MoE ViT-B/16 models, and ExFusion-mb ViT-B/16 on the ImageNet-1k dataset.}
\begin{tabular}{c|cccccc}
\toprule
\textbf{Method} &\textbf{ Top-1(\%)} & \textbf{\#Params} & \textbf{Storage} & \textbf{Memory} & \textbf{GPU Hours} \\
\midrule
Dense ViT-B    &77.9 &86.6M   &330MB  &19,201MB &240h \\
MoE-8-1 ViT-B  &71.5 &485.6M  &1.8GB  &29,475MB &664h \\
MoE-16-1 ViT-B &75.3 &939.0M  &3.5GB  &30,619MB &672h \\
MoE-32-1 ViT-B &75.6 &1845.8M &6.9GB  &32,866MB &696h \\
MoE-64-1 ViT-B &75.8 &3659.3M &13.7GB &37,380MB &720h \\
\textbf{ExFusion-mb ViT-B}  &\textbf{81.1} &\textbf{86.6M} &\textbf{330MB} &\textbf{25,327MB} &\textbf{264h} \\
\bottomrule
\end{tabular}
\label{tab:im1k-vit}
\end{table}

\begin{table}[!t]
\centering
\setlength\tabcolsep{15pt}
\caption{Results on the ImageNet-1k.}
\label{tab:im1k}
\begin{tabular}{@{}c|cc|c@{}}
\toprule
\textbf{Method} & \textbf{\#Params} & \textbf{FLOPs} & \textbf{Top-1 (\%)} \\ 
\hline\hline
ViT-S  & \multirow{2}{*}{22.1M}  & \multirow{2}{*}{4.6G}    & 71.8   \\ 
\textbf{ViT-S$\dagger$}  &   &   & \textbf{80.4(+8.6)}   \\ \hline
ViT-B  & \multirow{2}{*}{86.6M}   & \multirow{2}{*}{17.6G}   &  77.9 \\ 
\textbf{ViT-B$\dagger$}  &   &   &  \textbf{81.1(+3.2)} \\  \hline\hline
DeiT-T  & \multirow{2}{*}{5.7M}  & \multirow{2}{*}{1.2G}   &  72.2 \\
\textbf{DeiT-T$\dagger$}  &   &   &  \textbf{74.1(+1.9)} \\ \hline
DeiT-S & \multirow{2}{*}{22.1M}  & \multirow{2}{*}{4.6G}   &  79.8 \\
\textbf{DeiT-S$\dagger$}  &   &   &  \textbf{80.6(+0.8)} \\ \hline\hline
PVT-T  & \multirow{2}{*}{13.2M}  & \multirow{2}{*}{1.9G}   &  75.1 \\
\textbf{PVT-T$\dagger$}  &   &   &  \textbf{75.4(+0.3)} \\ \hline
PVT-S  & \multirow{2}{*}{24.5M}  & \multirow{2}{*}{3.8G}   &  79.8 \\
\textbf{PVT-S$\dagger$}  &   &   &  \textbf{80.7(+0.9)} \\ \hline
PVT-M  & \multirow{2}{*}{44.2M}  & \multirow{2}{*}{6.7G}   & 81.2  \\
\textbf{PVT-M$\dagger$}  &   &   &  \textbf{82.1(+0.9)} \\ \hline
PVT-L  & \multirow{2}{*}{61.4M}  & \multirow{2}{*}{9.8G}   & 81.7  \\
\textbf{PVT-L$\dagger$}  &   &   &  \textbf{82.5(+0.8)} \\ \hline\hline
Swin-T  & \multirow{2}{*}{29M}  & \multirow{2}{*}{4.5G}   &  81.3 \\
\textbf{Swin-T$\dagger$}  &   &   & \textbf{ 81.5(+0.2)} \\ \hline
Swin-S  & \multirow{2}{*}{50M}  & \multirow{2}{*}{8.7G}   &  83.0 \\
\textbf{Swin-S$\dagger$}  &   &   &  \textbf{83.2(+0.2)} \\ \hline
Swin-B  & \multirow{2}{*}{88M}  & \multirow{2}{*}{15.4G}   &  83.5 \\
\textbf{Swin-B$\dagger$}  &   &   &  \textbf{83.7(+0.2)} \\
\bottomrule
\end{tabular}
\end{table}

\subsection{Image Classification}
\label{sec:imnet-1k}

ImageNet-1k \cite{imagenet1k} is a widely used image classification dataset for visual model pre-training, which contains 1.28M images for training and 50K images for validation.  To ensure a fair comparison, we utilize the same training recipe as the baseline models.
Specifically, AdamW \cite{adamw} is utilized as the optimizer to train all models from scratch for 300 epochs, and weight decay is set to 0.05. We introduce a linear warmup of 20 epochs and use cosine decay to adjust the learning rate. The initial learning rate is established at 1e-3 for a batch size of 1,024, scaling linearly with the batch size.
For data augmentation techniques, we follow the previous schemes and also employ RandAugment \cite{randaugment}, Mixup \cite{mixup}, CutMix \cite{cutmix}, and random erasing \cite{randomerasing}.

Firstly, to verify the efficiency of our method, we pre-train the ViT-B/16 dense model, along with its corresponding MoE and ExFusion-mb models. As shown in Table \ref{tab:im1k-vit} and Figure \ref{fig:head}(b), compared to the Vanilla MoE models, ExFusion-mb not only achieves the best pre-training performance but also requires the lowest training overhead. Surprisingly, our method only increases the training duration by 10\% compared to the dense model.

\begin{table}[!t]
\centering
\renewcommand\arraystretch{1.2}
\setlength\tabcolsep{15pt}
\caption{Results on the ADE20K.}
\label{tab:ade20k}
\begin{tabular}{c|cc|c}
\toprule
\multicolumn{4}{ c }{\textbf{(a) Semantic FPN}} \\
\hline
\textbf{Backbone} &\textbf{Iter.} & \textbf{\#Params} & \textbf{mIoU (\%)} \\
\hline
PVT-T  & \multirow{2}{*}{40K} & \multirow{2}{*}{17M} & 35.7 \\
\textbf{PVT-T$\dagger$}  &  & &\textbf{37.2}  \\ \hline
PVT-S & \multirow{2}{*}{40K}  & \multirow{2}{*}{28M} & 39.8 \\
\textbf{PVT-S$\dagger$}  &  & &\textbf{42.0}  \\ \hline
PVT-M & \multirow{2}{*}{40K}   & \multirow{2}{*}{48M} & 41.6 \\
\textbf{PVT-M$\dagger$} &  & &\textbf{43.6} \\ \hline
PVT-L& \multirow{2}{*}{40K}   & \multirow{2}{*}{65M} & 42.1 \\
\textbf{PVT-L$\dagger$}  &  & &\textbf{44.4}  \\
\hline
\multicolumn{4}{ c }{\textbf{(b) UperNet}} \\
\hline
\textbf{Backbone} & \textbf{Iter.} & \textbf{\#Params} & \textbf{mIoU (\%)} \\
\hline
DeiT-T  & \multirow{2}{*}{80K}   & \multirow{2}{*}{34M} &38.0  \\
\textbf{DeiT-T$\dagger$}   &  & &\textbf{40.4}  \\ \hline
DeiT-T  & \multirow{2}{*}{160K}   & \multirow{2}{*}{34M} &  38.7\\
\textbf{DeiT-T$\dagger$}  &   & &\textbf{40.7}  \\ \hline
DeiT-S  & \multirow{2}{*}{80K}  & \multirow{2}{*}{52M} & 43.0 \\
\textbf{DeiT-S$\dagger$}   &  & &\textbf{43.5}  \\ \hline
DeiT-S  & \multirow{2}{*}{160K}   & \multirow{2}{*}{52M} &43.8  \\
\textbf{DeiT-S$\dagger$}   &  & &\textbf{44.8}  \\ \bottomrule
\end{tabular}
\end{table}

\begin{table}[!t]
\centering
\setlength\tabcolsep{10pt}
\caption{Comparison results on the COCO dataset.}
\label{tab:coco}
\centering
\begin{tabular}{ c|c|ccc }
\toprule
\multicolumn{5}{c}{\textbf{(a) Mask R-CNN}} \\
\midrule
\textbf{Backbone} & \textbf{\#Params} & \textbf{$\textup{AP}^b$} & \textbf{$\textup{AP}^b_{50}$} & \textbf{$\textup{AP}^b_{75}$} \\
\hline
PVT-M & \multirow{2}{*}{63.9M} & 42.0 & 64.4 & 45.6 \\
\textbf{PVT-M$\dagger$} & & \textbf{42.6} & \textbf{65.1} & \textbf{46.4} \\ \hline
PVT-L & \multirow{2}{*}{81.0M} & 42.9 & 65.0 & 46.6 \\
\textbf{PVT-L$\dagger$} & & \textbf{43.0} & \textbf{65.2} & \textbf{46.8} \\
\bottomrule
\end{tabular}
\end{table}

\begin{table}[!t]
\centering
\setlength\tabcolsep{10pt}
\begin{tabular}{ c|c|ccc }
\toprule
\multicolumn{5}{c}{\textbf{(b) RetinaNet}} \\
\midrule
\textbf{Backbone} & \textbf{\#Params} & \textbf{AP} & \textbf{$\textup{AP}_{50}$} & \textbf{$\textup{AP}_{75}$} \\
\hline
PVT-M & \multirow{2}{*}{53.9M} & 41.9 & 63.1 & 44.3 \\
\textbf{PVT-M$\dagger$} & & \textbf{42.6} & \textbf{63.7} & \textbf{45.5} \\ \hline
PVT-L & \multirow{2}{*}{71.1M} & 42.6 & 63.7 & 45.4 \\
\textbf{PVT-L$\dagger$} & & \textbf{42.9} & \textbf{64.2} & \textbf{45.6} \\
\bottomrule
\end{tabular}
\end{table}

Furthermore, we validate the efficacy of our method across four classic Transformer-based models (ViT \cite{vit}, DeiT \cite{deit}, PVT \cite{pvt}, Swin \cite{swin}), and report the Top-1 accuracy. As illustrated in Table \ref{tab:im1k}, training existing Transformer-based models based on ExFusion-mb framework consistently yields performance improvements without increasing the number of parameters or computational cost. For example, our ViT-B$\dagger$ and DeiT-T$\dagger$ achieve gains of 3.2\% and 1.9\%, respectively. Moreover, our PVT-M$\dagger$ surpasses PVT-L by 0.4\%, using only 72\% of the parameters and 68\% of the computational cost. These experimental results verify the strong advantage and applicability of our method across a variety of Transformer-based models.

\subsection{Semantic Segmentation}
The ADE20K dataset \cite{ade20k} is a predominant benchmark for semantic segmentation, consisting of 20K training images and 2K validation images.
Upon pre-training PVT and DeiT models on the ImageNet-1k dataset, we further fine-tune these models on ADE20K. 
The performance of the pre-trained model is assessed through two segmentation methods, namely Semantic FPN \cite{sfpn} and UperNet \cite{upernet}.
As demonstrated in Table \ref{tab:ade20k}, the pre-trained models obtained through ExFusion-mb consistently achieve performance improvements in segmentation tasks.
More precisely, our method achieves an average mIoU improvement of 1.7\%, while preserving the same parameter count.

\subsection{Object Detection}

COCO 2017 \cite{cocodataset} is a widely recognized dataset for object detection, comprising 118K training images and 5K images for validation. We evaluate the performance of pre-trained models using two detection methods, Mask R-CNN \cite{maskrcnn} and RetinaNet \cite{retinanet}. As shown in Table \ref{tab:coco}, employing ExFusion-mb for enhanced pre-training of PVT models can improve the performance in detection tasks. 
Notably, within the RetinaNet framework, our PVT-M$\dagger$ outperforms PVT-L by 0.1\% $\textup{AP}_{75}$ with only 75\% of parameters.

\subsection{Natural Language Understanding}

\begin{figure}[!t]
    \centering
    \includegraphics[width=0.98\linewidth]{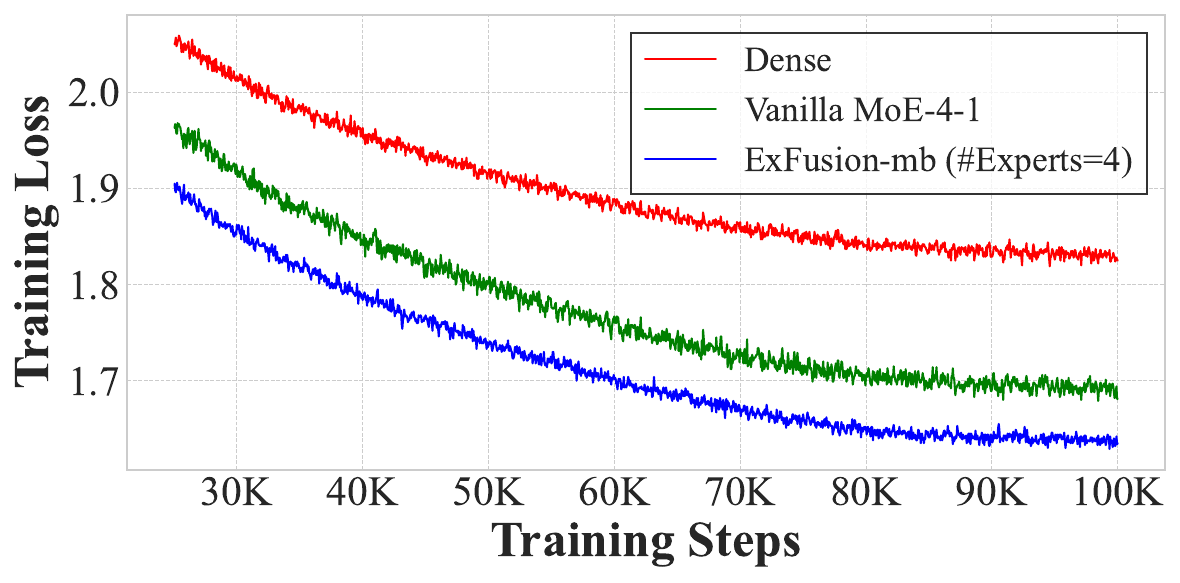}
    \caption{Comparison results on C4.}
    \label{fig:t5}
\end{figure}

\begin{table}[!t]
    \centering
    \caption{Fine-tuning results on GLUE.}
    \label{tab:t5}
    \scriptsize
    \renewcommand{\arraystretch}{1.2} 
    \begin{tabular}{c|cccccc}
        \toprule
        \textbf{Method} & \textbf{\rotatebox{90}{SST-2}} & \textbf{\rotatebox{90}{MRPC}} & \textbf{\rotatebox{90}{QQP}} & \textbf{\rotatebox{90}{MNLI}} & \textbf{\rotatebox{90}{QNLI}} & \textbf{\rotatebox{90}{Avg.}} \\
        \midrule
        Dense                & 91.3 & 86.8 & 90.7 & 83.5 & 90.2 & 88.5 \\
        MoE-4-1              & 93.1 & 89.2 & 90.4 & 84.0 & 91.3 & 89.6 \\
        \textbf{ExFusion-mb} & \textbf{93.7} & \textbf{90.6} & \textbf{91.0} & \textbf{86.3} & \textbf{91.9} & \textbf{90.7} \\
        \bottomrule
    \end{tabular}
\end{table}

To verify the versatility of the ExFusion framework, we also conduct experiments on natural language understanding tasks.
We first pre-train T5 model~\cite{t5-c4} on the large-scale textual Colossal Clean Crawled Corpus (C4) corpus.
We employ the nanoT5 project \cite{nanot5} to conduct T5-base model pre-training.
Specifically, we train the Dense T5-base, MoE-4-1 T5-base, and T5-base with ExFusion-mb on the C4 dataset for 100K steps, with an initial learning rate of 1e-2 and a batch size of 2,048, while all other settings remained consistent with the nanoT5 project.
Figure~\ref{fig:t5} shows the pre-training loss variations of the above three models.
Surprisingly, our method not only achieves the best performance but also exhibits a faster convergence rate.
For instance, our method reaches the same level of convergence at 30K steps as the dense model does at 100K steps, which is 3.3x faster. Furthermore, it is noteworthy that the GPU hours required for ExFusion-mb are only 1/3 of those for MoE-4-1 (644 hours vs. 1,742 hours).
Finally, we utilize the GLUE benchmark\footnote{GLUE is a benchmark designed to evaluate the performance of models on a variety of natural language understanding (NLU) tasks. We utilize 5 tasks within GLUE for evaluating the fine-tuning performance of pre-trained models.} to verify the pre-training effect of the T5-base models. As shown in Table \ref{tab:t5}, the T5-base model pre-trained via ExFusion-mb also achieves exceptional downstream performance.

\subsection{Ablation Studies}

\begin{table}[!t]
    \centering  
    \begin{minipage}{0.48\linewidth}
      \centering
      \captionof{table}{Results on different ExFusion variants.}
      \label{tab:vom}
      \begin{tabular}{c|c}
      \toprule
        Framework & Acc. \\
        \midrule
        baseline & 72.20 \\
        ExFusion-sw &  73.71 \\
        ExFusion-dw &  73.96 \\
        ExFusion-mb &  74.10 \\ 
        \bottomrule
      \end{tabular}
    \end{minipage}%
    \hfill
    \begin{minipage}{0.48\linewidth}
      \centering
      \captionof{table}{Results on the number of experts.}
      \label{tab:noe}
      \begin{tabular}{c|c}
      \toprule
        \#Experts & Acc. \\
        \midrule
        4 & 74.10\\
        8 & 73.56\\
        16 & 73.28\\
        \bottomrule
      \end{tabular}
    \end{minipage}
    \end{table}
    
\begin{table}[!t]
    \begin{minipage}{0.48\linewidth}
      \centering
      \captionof{table}{Results on the momentum value.}
      \label{tab:mv}
      \begin{tabular}{c|c}
      \toprule
        Value of $\delta$ & Acc. \\
        \midrule
        0.90 & 73.72 \\
        0.92 & 74.06 \\
        0.95 & 74.10 \\
        0.98 & 74.08 \\
        \bottomrule
      \end{tabular}
    \end{minipage}%
    \hfill
    \begin{minipage}{0.48\linewidth}
      \centering
      \captionof{table}{Results on router settings.}
      \label{tab:gr}
      \begin{tabular}{cc|c}
      \toprule
        Router & \#Params & Acc. \\
        \midrule
        w Shared  & 9.2K & 74.10 \\
        w/o Shared & 46.1K & 73.95 \\
        \bottomrule
      \end{tabular}
    \end{minipage}
    \end{table}

In this section, we validate the effectiveness of each component within the ExFusion framework via comprehensive ablation studies. Unless specifically noted, we utilize DeiT-T as the baseline model, employ ExFusion-mb with four experts as the training strategy, and conduct experiments on the ImageNet-1k dataset.

\subsubsection{Variants of ExFusion.}

In the aforementioned experiments, we present the results for the best-performing variant, ExFusion-mb. Therefore, as shown in Table \ref{tab:vom}, we verify the effectiveness of the other two variants. The results indicate that without considering data-dependent information, both the static-weight-based ExFusion-sw and the dynamic-weight-based ExFusion-dw achieve suboptimal performance. However, compared to the baseline, they still achieve gains of 1.51\% and 1.76\%, respectively.

\subsubsection{Number of Experts.}

In MoE models, the number of experts is an important factor that affects the performance. To investigate the impact of this factor, we evaluate the ExFusion-mb method with varying numbers of experts. The results in Table \ref{tab:noe} indicate that increasing the number of experts actually leads to a performance decline. One plausible explanation could be that, with the total number of training epochs held constant, increasing the number of experts may result in insufficient training for each expert. Therefore, we have fixed the number of experts to 4 for ExFusion-mb.

\subsubsection{Momentum Values.}

In ExFusion-mb, a momentum update mechanism is introduced to enable the comprehensive consideration of both current and historical weights. In this context, a larger momentum value \(\delta\) indicates a lower influence of the current weights, while a smaller value suggests a reduced impact of the historical weights. The ablation results for \(\delta\), as shown in Table \ref{tab:mv}, leads us to fix the momentum value at 0.95.

\subsubsection{Router Settings.}

ExFusion-mb employs a shared router and a memory bank to regulate the weights of experts. We also investigate the effects of a non-sharing configuration. The results in Table \ref{tab:gr} show that the adoption of a sharing mechanism not only achieves superior performance but also reduces the number of router parameters by 80\%.

\subsubsection{Placement of MoE Layers.}

In the MoE model, the placement of MoE layers is crucial.
To further verify the importance of MoE layers, we conduct ablation studies to find the best positions for replacing dense FFN layers with multi-experts layers in Transformer.
According to the results in Table \ref{tab:rp}, ExFsion-mb achieves the best performance when all FFN layers are replaced.

\subsubsection{Training Duration.}

As illustrated in Table \ref{tab:ts}, employing four experts for pre-training in ExFusion-mb results in an increase of only 8\% in training time. Notably, even when expanding to 128 experts, the increase in training time is limited to only 11\%. Additionally, as shown in Table \ref{tab:im1k-vit}, compared to the vanilla MoE, ExFusion-mb significantly reduces both training time and memory usage for the same activated parameters. Furthermore, compared to dense models, ExFusion-mb results in only a 10\% increase in training time, while maintaining identical inference costs. This indicates that our proposed ExFusion-mb method does not substantially increase the training time as the number of experts scales.

\subsection{Visualization}

\begin{figure*}[!t]
    \centering
    \includegraphics[width=1.00\textwidth]{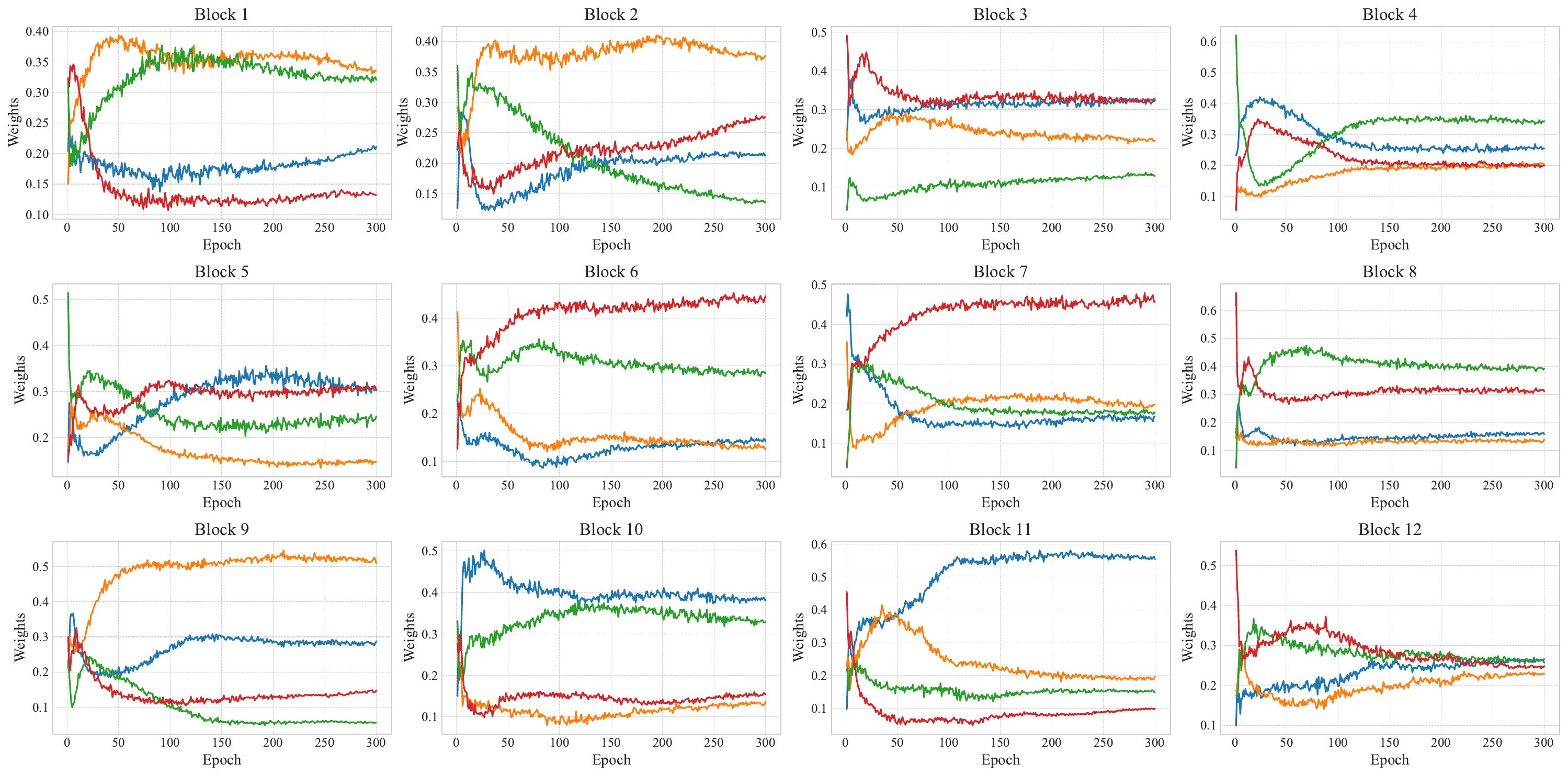}
    \caption{Visualization of expert weight variations in DeiT-T$\dagger$.}
    \label{fig:vis}
\end{figure*}

To verify the trend of weight updates in the ExFusion-mb, we visualize the change of expert weights in DeiT-T$\dagger$. As shown in Figure \ref{fig:vis}, the four lines in each subplot respectively represent the weight variation of four experts. The results indicate that ExFusion-mb can effectively adjust the importance of each expert during the training process, achieving robust convergence. Furthermore, the differences among the weights further verify that experts are specialized with routing preferences, which is consistent with the consensus of many studies on MoE \cite{understandingmoe,llama-moe-2023}.

\begin{table}[!t]    
    \begin{minipage}{0.48\linewidth}
      \centering
      \captionof{table}{Results on replacement layers.}
      \label{tab:rp}
      \begin{tabular}{c|c}
      \toprule
        Replacement layers &  Acc.\\
        \midrule
        \{6, 7, 8, 9, 10, 11\} & 72.98\\
        \{0, 2, 4, 6, 8, 10\} & 74.07\\
        \{1, 3, 5, 7, 9, 11\} & 73.29\\
         All layers & 74.10\\
        \bottomrule
      \end{tabular}
    \end{minipage}%
    \hfill
    \begin{minipage}{0.48\linewidth}
      \centering
      \captionof{table}{Results on training duration.}
      \label{tab:ts}
      \begin{tabular}{c|c}
      \toprule
        \#Experts & Training Time\\
        \midrule
        -  & ×1 \\
        4  & ×1.02\\
        8  &×1.08\\
        128 & ×1.11 \\
        \bottomrule
      \end{tabular}
    \end{minipage}
\end{table}

\section{Discussion and Limitation}

In this section, we provide further insights into the distinctions between the proposed ExFusion method and other advanced Mixture-of-Experts (MoE) models, which aim to enhance routing strategies, specialize expert architectures, or customize MoE for specific tasks. We also address the limitations of our approach and discuss areas requiring further exploration.

\subsection{Efficiency and Deployment-Friendliness}

A key advantage of ExFusion lies in its efficiency and deployment-friendliness, particularly during the inference stage. Advanced MoE models, even if they activate only a small subset of experts, still require storing and managing all experts and routers during inference. This leads to significant increases in memory and storage requirements as the number of experts grows. In contrast, ExFusion fuses all experts into a single Feed-Forward Network (FFN) after training, resulting in an inference-time architecture that is identical to the dense Transformer model, both in terms of structure and computational complexity. This characteristic of being "large during training, compact during inference" provides distinct advantages in resource-constrained environments, such as mobile devices, multimedia terminals, and cloud-edge collaboration systems commonly found in Telecommunications, Multimedia, and Mobile (TMM) applications.

\subsection{Compatibility with Existing Dense Transformer Ecosystems}

ExFusion is highly compatible with existing dense Transformer ecosystems. The output model of ExFusion can seamlessly replace any Transformer backbone without necessitating changes to downstream pipelines used for tasks such as object detection, semantic segmentation, or information retrieval. In contrast, many advanced MoE models require specialized support and custom engineering implementations for each downstream task. This flexibility makes ExFusion particularly suitable for industrial multimedia systems where adaptability and ease of integration are paramount.

\subsection{Limitations of ExFusion}

Despite its promising performance in efficient Transformer training, there are several limitations to the ExFusion method that should be considered:

\subsubsection{Limitations with Large Numbers of Experts}

ExFusion fuses all experts into a single expert during training. When the number of experts becomes extremely large, each expert might exhibit high diversity, and excessive averaging could result in overly strong ``smoothing". This can hinder the preservation of specialized behaviors. As seen in Table VII, performance declines when the number of experts increases from 4 to 16, highlighting the trade-off between expert diversity and effective knowledge fusion.

\subsubsection{Specialized Expert Behaviors and Inference-Stage Routing}

For certain tasks, such as multi-task learning or multimodal tasks, the model benefits from retaining distinct experts during inference. These tasks often rely on a router for conditional computation, enabling the model to fully utilize the specialized capabilities of each expert. In such cases, preserving the MoE structure during inference may be more advantageous, whereas ExFusion’s fusion of experts could potentially lose the benefits of specialization.

\subsubsection{Insufficient Validation on Extremely Large-Scale Models and Datasets}

{The current experiments primarily focus on models such as ViT-B, Swin-B, and T5-base. For models with hundreds of billions of parameters—such as large language models or extremely large vision models—the behavior of training and fusion still requires further systematic investigation. These challenges warrant further research before ExFusion can be fully deployed in large-scale scenarios.

\subsection{Distinction from Other Methods}

To further clarify the uniqueness of ExFusion, we compare it with various related methods.

\subsubsection{Comparison with MoE Models}

ExFusion introduces a multi-expert structure only during training, and ultimately merges it into a single-expert model via weight fusion. During inference, ExFusion’s parameter quantity and computational complexity are strictly equivalent to those of a dense Transformer model. This contrasts sharply with MoE models such as CBDMoE~\cite{tmm_moe4}, which enhance the MoE structure but still retain multiple experts during inference, resulting in higher storage and computation costs.

\subsubsection{Comparison with Efficient Tuning Methods for Transformer}

While methods like PEFT \cite{convpass,nlpadapter,ruan2024understanding,ruan2025tte,gist,idat,nlplora,peftreview} focus on optimizing the pre-existing architecture to improve efficiency during training, ExFusion targets efficient expansion and compression during the pre-training stage. It leverages a multi-expert structure to enhance the backbone model’s capability without altering the final architecture. As a result, ExFusion complements, rather than competes with, efficient training methods, particularly in the context of scalable training and model efficiency.

\subsubsection{Comparison with Efficient MoE Training Methods}

When compared to efficient MoE training methods such as Residual Mixture of Experts (RMoE)~\cite{wu2022residual} and Weight-Ensembling MoE (WEMoE)~\cite{shen2025efficient}, ExFusion stands out due to its zero-inference overhead. RMoE reduces training costs by factorizing weights but still requires MoE models with sparse computation support. WEMoE merges fine-tuned models but retains a dynamic routing mechanism during inference. In contrast, ExFusion uses the multi-expert structure solely as a temporary training scaffold and collapses all experts into a single dense network after training. This results in a final model that is mathematically identical to a standard Transformer, eliminating the complexities associated with sparse operations or dynamic routing during deployment.

\section{Conclusion}
\label{sec5}

In this paper, we introduce ExFusion, which involves the weighted fusion of multiple experts within a Mixture-of-Experts model during and after training, aimed at enhancing the performance of Transformer-based models without increasing the overhead. We have developed three variants of the ExFusion based on static, dynamic, and memory bank strategies to improve the flexibility of our approach. Throughout extensive experiments across a variety of tasks, including image classification, semantic segmentation, object detection, and language understanding, we demonstrate that our method significantly improves the performance of various Transformer-based models without additional parameters or computational cost, showcasing its strong universality and applicability. For future work, we plan to extend ExFusion to larger-scale models, as this direction remains unexplored in the current study. Another promising avenue is to adapt ExFusion for multi-sensor or multi-modal fusion scenarios, such as the joint modeling of radar, LiDAR, and camera data in autonomous driving systems, which we also reserve for subsequent investigations.

\section*{Acknowledgments}
This work was partially supported by the National Natural Science Foundation of China under Grant No.62301315, Startup Fund for Young Faculty at SJTU (SFYF at SJTU) under Grant No.23X010501967 and Shanghai Municipal Health Commission Health Industry Clinical Research Special Project under Grant No.202340010.
The authors would like to thank the anonymous reviewers for their valuable suggestions and constructive criticisms.

\bibliographystyle{IEEEtran}


\bibliography{TCSVT}

\end{document}